\documentclass{article}
\usepackage{subfig}
\usepackage{amsmath}
\usepackage{booktabs}
\usepackage[pdftex]{graphicx}
\usepackage{epstopdf}
\usepackage{enumitem}
\usepackage{xcolor}

\usepackage{todonotes}

\begin{document}

\title{Systematic and Realistic Testing in Simulation of Control Code for Robots in Collaborative Human-Robot Interactions} 

\author{Dejanira Araiza-Illan, David Western, Anthony G. Pipe and Kerstin Eder\footnote{
Dejanira Araiza-Illan, David Western, and Kerstin Eder are with the Department of Computer Science and Bristol Robotics Laboratory,
University of Bristol, Bristol, UK. E-mail: \texttt{\{dejanira.araizaillan,david.western,kerstin.eder\}@bristol.ac.uk}.
Anthony Pipe is with the Faculty of Engineering Technology and Bristol Robotics Laboratory, University of the West of England, Bristol, UK. E-mail: \texttt{tony.pipe@brl.ac.uk}}}

\date{}

\maketitle

\begin{abstract}
Industries such as flexible manufacturing and home care will be transformed by the presence of robotic assistants.
Assurance of safety and functional soundness for these robotic systems will require rigorous verification and validation.
We propose testing in simulation using Coverage-Driven Verification (CDV) to guide the testing process in an automatic and systematic way. 
We use a two-tiered test generation approach, where abstract test sequences are computed first and then concretized (e.g., data and variables are instantiated), to reduce the complexity of the test generation problem. 
To demonstrate the effectiveness of our approach, we developed a testbench for robotic code, running in ROS-Gazebo, that implements an object handover as part of a human-robot interaction (HRI) task.
Tests are generated to stimulate the robot's code in a realistic manner, through stimulating the human, environment, sensors, and actuators in simulation. 
We compare the merits of unconstrained, constrained and model-based test generation in achieving thorough exploration of the code under test, and interesting combinations of human-robot interactions.
Our results show that CDV combined with systematic test generation achieves a very high degree of automation in simulation-based verification of control code for robots in HRI.
\end{abstract}

\section{Introduction}\label{sc:introduction}

Robotic assistants for industrial and domestic applications are designed to interact and collaborate directly with humans. 
These close interactions have ethical and legal implications.
Consequently, the safety and functional soundness of such technologies needs to be demonstrated, for them to become viable commercial products~\cite{ROMAN14}. 
Currently, a physical separation between robots and humans is enforced for safety, besides restrictions of speed and force.\footnote{Standards ISO~13482:2014 for robotic assistants and ISO~10218 (parts I and II) for industrial robotics.}
These restrictions limit the scope of the applications for collaborative robots.  
To demonstrate that speed and force restrictions are being met, and thus safety can
be assured even without physical separation, the software that controls these
robotic platforms must be subjected to rigorous verification and validation
(V\&V) processes.
Software V\&V needs to consider the robotic system as a whole entity, i.e.\ the software coupled with its hardware and electronics, as well as the reality and uncertainties of the target environments.

V\&V of human-robot interactions (HRI) is challenging. 
The robot's environment is dynamic and uncertain (e.g., it includes people). 
Current V\&V methods and tools are limited by computational resource bounds, restricting the degree of realism, detail, and exhaustiveness of exploration. 
Formal methods, e.g.\ model checking and theorem proving, are exhaustive and provide proof of requirement satisfaction, at the cost of employing highly abstracted models of the robotic systems and HRIs due to computational constraints, as in~\cite{stocker12verification,webster14formalshort}. 
Testing in simulations allows realism and detail~\cite{Petters2008,Pinho2014}, at the cost of not being exhaustive with respect to the possibilities in the system under test (SUT), nor providing guarantees of requirement satisfaction.

Available verification methodologies from other domains, such as the microelectronics design industry, provide systematic and targeted approaches to maximize ``coverage'' (i.e., the extent to which a system's design has been explored) in testing. 
One of these methodologies is Coverage-Driven Verification (CDV), where various coverage models are used to assess exploration of the SUT and V\&V completion~\cite{Piziali2004}. 
Tests that maximize coverage --i.e., effective tests-- are generated (mostly) automatically, coupled with feedback loops (automatic or manual) from automated coverage metrics collection, and automatic checks of (mostly) the SUT's response.

In test generation, constraints are commonly employed to bias testing towards rare events for coverage closure, after applying pseudo-random approaches to achieve exploration of the SUT~\cite{Piziali2004,Gaudel2011}. 
Model-based test generation uses formal methods (e.g., model checking) or other techniques to explore models in order to obtain test bias~\cite{Utting2012}. 
Nonetheless, computing tests that stimulate robotic code in a realistic or human-like manner, as it would happen in a real-life HRI scenario, makes the test generation problem quite complex. 

We manage complexity via a two-tiered test generation approach. Abstract test sequences are generated first, and then instantiated to obtain concrete tests that stimulate the robotic code indirectly --i.e., the tests stimulate the human, environment, sensors and actuators in simulation, these then stimulate the robot.
For example, a test requires a human to send voice commands to activate the robot in a particular order, expressed as `send voice command' actions in the abstract layer.
Code that executes these `human' actions is assembled according to the test action sequences. 
The concretization of these action sequences is the production of timed sequences from the human voice model in simulation, that will stimulate simulated voice sensors, and then will send their readings to the robot's code to stimulate it. 
This two-tiered process is employed in model-based testing~\cite{Utting2012}. 
In this paper we apply unconstrained, constrained, and model-based abstract test generation, coupled with test concretization via uniform sampling from classified ranges for variables and parameters. 
We demonstrate the complementary strengths of exploratory and targeted tests, particularly through model-based test generation, in achieving high levels of coverage for different coverage models, including code, cross-product, and assertions (requirements).

We tested the code for an object handover interaction between a humanoid torso and a person, envisaged for cooperative manufacture tasks, in a simulator developed in Robot Operating System\footnote{http://www.ros.org/} (ROS) and Gazebo\footnote{http://gazebosim.org/}, a 3D physics simulator. 
We employed a CDV testbench prototype developed for our simulator, fully compatible with ROS-Gazebo\footnote{Available at: https://github.com/robosafe/testbench/v3}. 
This paper extends our previous work in~\cite{CDV2015}, with more requirements, coverage models, generated tests and simulation runs. 
Our testbench prototype is transferable and extendible to other robotic simulators based on ROS, and other collaborative and assistive applications.

The paper is structured as follows. We present the handover scenario in Section~\ref{sc:casestudy}. The testbench components are presented in Section~\ref{sc:CDV}. A discussion of V\&V and coverage results is presented in Section~\ref{sc:results}. Related work is presented in Section~\ref{sc:relatedwork}, and Section~\ref{sc:conclusions} concludes with an outlook on future work.

\section{Case Study: Robot to Human Object Handover Task}\label{sc:casestudy}

The object handover case study was chosen because it is critical in many HRI tasks, such as cooperative manufacture, or home care. 
The robot platform, BERT2, is a humanoid torso with two arms~\cite{lenz2010bert2}. 
A handover starts with voice activation from the person to the robot. 
The robot proceeds to pick up an object, holds it out to the human, and signals for the human to take it. 
The human indicates readiness to take the object through another voice command. 
Then, the robot will collect three sensor readings: ``pressure,'' indicating whether the human is holding the object (applying force against the robot's hold of the object); ``location,'' visually tracking that the person's hand is close to the object; and ``gaze,'' visually tracking that the person's head is directed towards the object. 
Each sensor reading is classified into $G\!=\!P\!=\!L\!=\!\{\bar{1},1\}$, where $1$ indicates the sensing was positive that the human is ready to receive the object, and $\bar{1}$ is any other sensing outcome, including null. 
After the sensing, the robot should decide to release the object if the human is ready, i.e.\ $GPL = (1,1,1)$ from the Cartesian product of the sensor readings (GPL for short), or it should decide not to release the object otherwise, i.e.\ $GPL \in \{ (\bar{1},*,*), (*,\bar{1},*), (*,*,\bar{1}) \}$, where $* \in \{1,\bar{1}\}$, within a time threshold. 
The person may disengage from the task before the robot makes a decision. The robot can time out whilst sensing, or while waiting for a signal.

A ROS `node' contains the robot's action control code, comprising 212 statements in Python. 
The code was structured as a FSM using the SMACH modules~\cite{SMACH}, to facilitate computing a model of it for model-based test generation.

\subsection{Requirements List}\label{ssc:requirements}

The following safety and functional requirements need to be verified, derived from the standard ISO~13482:2014 and previous work on handover interaction protocols and their testing in~\cite{Grigore2011,CDV2015}: 
\begin{enumerate}
\item If the gaze, pressure and location are sensed as correct, then the object shall be released. 
\item If the gaze, pressure or location are sensed as incorrect, then the object shall not be released. 
\item The robot shall make a decision before a threshold of time. 
\item The robot shall always either time out, decide to release the object, or decide not to release the object.
\item The robot shall not close the gripper when the human is too close.
\item The robot shall start in restricted speed.
\item The robot shall not collide with itself at high speeds.
\item The robot shall operate within allowable maximum values to avoid dangerous unintentional collisions with humans and other safety-related objects.
\end{enumerate}

The last requirement was implemented in four different manners, considering a speed threshold of 250 mm/s based on standard ISO~10218-1:2011:
\let\oldtheenumi=\theenumi
\renewcommand{\theenumi}{8\alph{enumi}}
\begin{enumerate}
\item The robot hand speed is always less than 250 mm/s.
\item If the robot is within 10 cm of the human, the robot's hand speed is less than 250 mm/s.
\item If the robot collides with anything, the robot's hand speed is less than 250 mm/s.
\item If the robot collides with the human, the robot's hand speed is less than 250 mm/s.
\end{enumerate}
\renewcommand{\theenumi}{\oldtheenumi}

\subsection{Handover Simulator}\label{ssc:simulator}

\begin{figure}[t]
\centering
\includegraphics[width=0.4\textwidth]{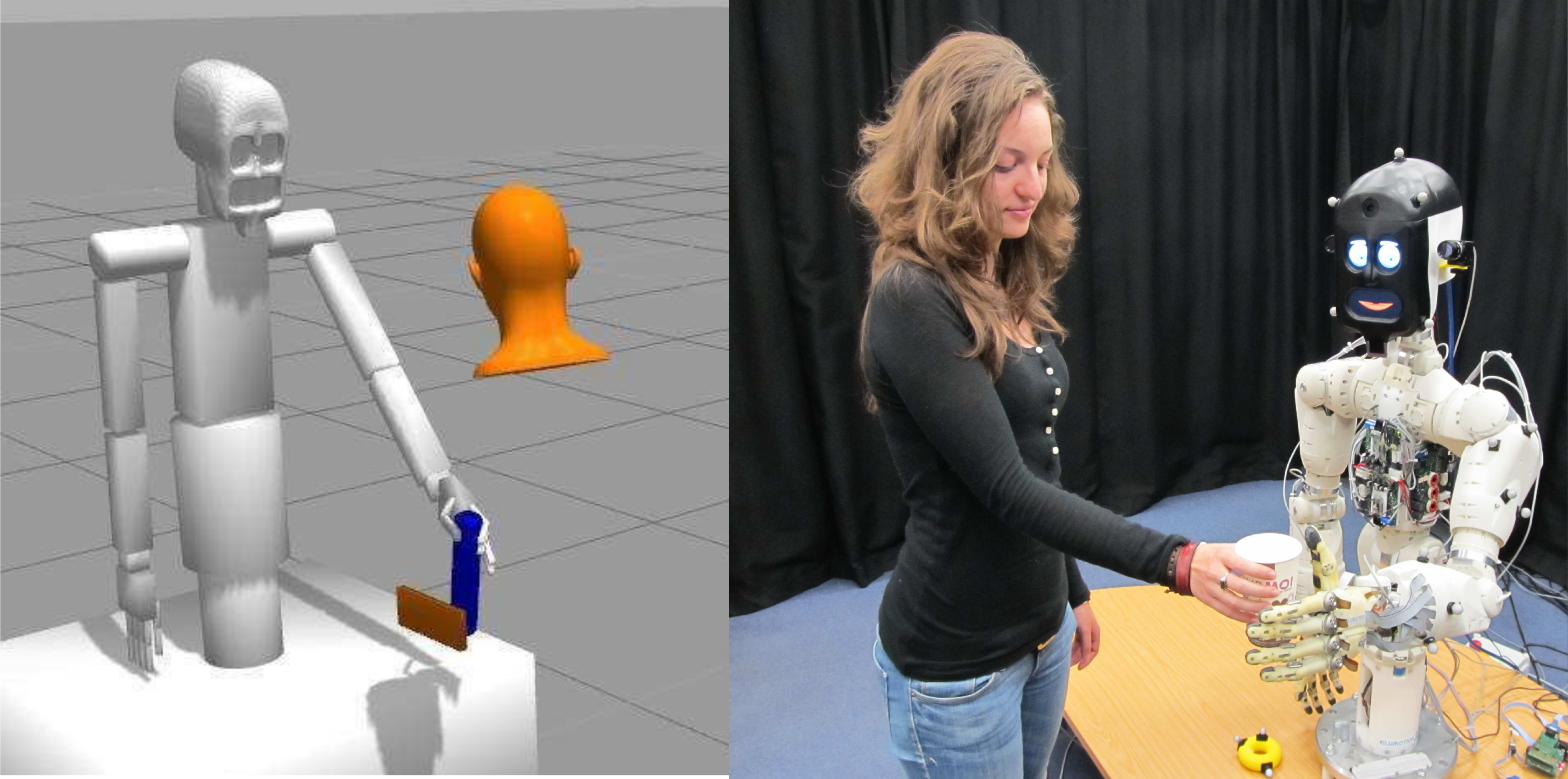}
\caption{The ROS-Gazebo simulation of a real handover.}
\label{Simulatorphoto}
\end{figure}

A simulator of the handover scenario was developed in ROS-Gazebo. 
ROS is an open-source platform for the development and deployment of robotics code, using C++ and/or Python. 
Gazebo is a 3D physics simulator, compatible with ROS. 
BERT2, a cylindrical object, and the person's head and hand were modelled in Gazebo, as shown in Fig.~\ref{Simulatorphoto}. 
Models were developed in code for the sensors and the human action enactment. 

\section{A CDV Testbench for a ROS-GAZEBO Simulator} \label{sc:CDV}

\begin{figure}[t]
\centering
\includegraphics[width=\textwidth]{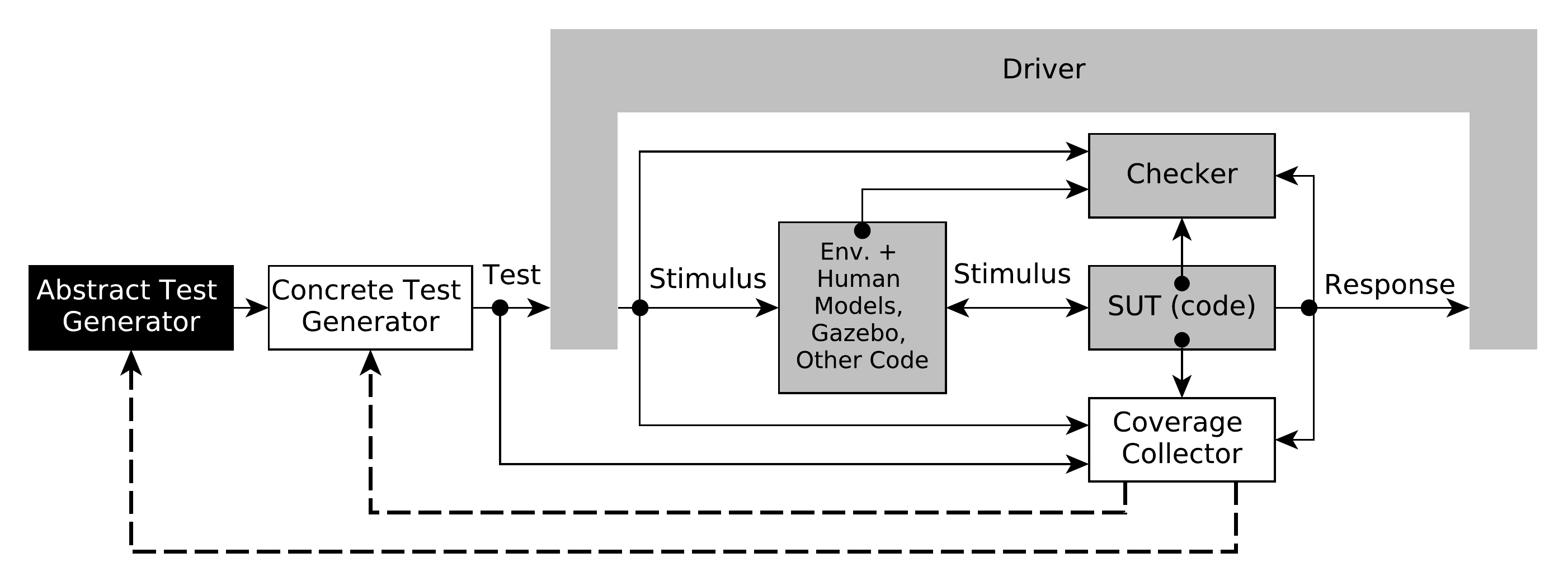}
\caption{Testbench and simulator elements in ROS-Gazebo}
\label{simstructure}
\end{figure}

In the CDV methodology, a verification plan indicates the requirements to test, and the coverage models and metrics to use over the SUT~\cite{Piziali2004,CDV2015}. 
A CDV testbench has four components: the {\bf Test Generator}, the {\bf Driver}, the {\bf Checker} and the {\bf Coverage Collector}. 
Figure~\ref{simstructure} shows our testbench, considering the ROS-Gazebo simulator's components. 
The simulator's design ensures the access to internal parameters in the robot's code and data about the physical models from Gazebo, to facilitate checking and coverage collection. 
The dotted line indicates feedback to the test generation for coverage closure and verification completion that may require human input.

\subsection{Test Generator}\label{sc:testgen}

The aim of the test generation process is to trigger bugs in the SUT (the robot's code), while exploring a wide range of scenarios. 
Guidance to produce effective tests comes from coverage and verification progress feedback. 
The generated tests must be valid and realistic, which makes the case for non-conventional software test generation approaches due to the complexity of ``stimulating the robot's code in a human-like manner''.

A test for the handover simulator is formed by an abstract test sequence for the human, environment, sensors and actuators (the environment surrounding the robotic code under test), which assembles code fragments to be executed concurrently by these simulator components. 
A concrete test is then computed, after parametrization, constraint solving and/or instantiation for all the individual parameters involved in the code fragments. 
We propose a two-tiered test generation approach to divide and simplify what would be a complex constraint solving, search or optimization problem. An abstract-to-concrete test construction is shown in Fig.~\ref{exampletest}. 

\begin{figure}[t]
\scriptsize
\centering
\renewcommand{\arraystretch}{1.2}
\begin{tabular}{r|ll|l}
\cline{2-3}
1&\verb+sendsignal+&\verb+activateRobot+ & Send human voice A1 for 5 sec.\\
2&\verb+setparam+ & \verb+time = 40+& Human waits $40 \times 0.05$ sec.\\
3&\verb+receivesignal+ & \verb+informHumanOfHandoverStart+ & Human waits for max. 60 sec.\\
4&\verb+sendsignal+ & \verb+humanIsReady+ &Send human voice A2 for 2 sec.\\
5&\verb+setparam+ & \verb+time = 10+ & Human waits $40 \times 0.05$ sec.\\
6&\verb+setparam+ & \verb+hgazeOk = true+ & Move human head in Gazebo to pose within ranges: \\
& & & offset $[0.1,0.2]$, distance $[0.5,0.6]$ and angle $[15,40)$ \\
\cline{2-3}
\end{tabular}
\caption{An abstract test sequence for the human to stimulate the robot's code (LHS), and its concretization from sampling from defined ranges (RHS).}
\label{exampletest}
\end{figure}

We explored three options for the abstract test generation: pseudorandom, constrained and model-based. 
Pseudorandom (for repeatability purposes) is, in principle, unconstrained with respect to any assumptions about the HRI protocol.
Thus, abstract test sequences are concatenated randomly, e.g., representing a person that disregards the handover protocol. 
To generate interesting tests, e.g, to verify a particular requirement, pseudorandom test generation can be biased using constraints. 
The implementation of these constraints requires significant manual input to be effective.
Model-based test generation techniques~\cite{Haedicke2012,Lakhotia2009,Utting2012} can target specific scenarios or requirements more effectively. 
In model-based test generation, a model of the system is explored or traversed in a systematic manner, e.g., through model checking for a requirement expressed as a temporal logic property~\cite{ClarkeMC}. 
A path through the model can be considered as a set of constraints~\cite{Lackner2012,Nielsen2003} for test generation. 

For model-based test generation, the model captures both the ideal robot's code functionality and the human/environment's actions, assuming both follow the handover protocol. % correctly.
We chose probabilistic-timed automata (PTA)~\cite{Hartmanns2009} models constructed manually in UPPAAL\footnote{http://www.uppaal.org/}, to capture uncertain actions such as disengaging from the task, and the important aspect of human-like response timing in HRI. 
Requirements 1 to 4 (Section~\ref{ssc:requirements}) were expressed as temporal logic properties, and model checked in UPPAAL. 
A witness trace (or path over the automata) is produced as a result of model checking, from which an abstract test sequence is extracted, disregarding the robot's actions in the trace.

\subsection{Driver}\label{sc:driver}

The Driver distributes the resulting concrete tests into the simulator components, to be enacted to stimulate the robot indirectly. 
The Driver reacts to the responses of the SUT if necessary (a ``reactive Driver''). 

\subsection{Checker}\label{sc:checker}

The Checker monitors the response of the SUT during simulation, to detect failures and bugs. 
Automata-based assertion monitors were implemented manually for all the requirements in Section~\ref{ssc:requirements}, as in~\cite{CDV2015}. 
Events can be monitored at different abstraction levels, from ``the robot received the correct command'' (abstract), to ``speed is less than the safe thresholds'' (semi-continuous signals or variables).
For example, the assertion monitor for Req. 5 is triggered every time the code executes the \verb+hand(close)+ function. 
The pose of the human hand is queried from the physical models in Gazebo. 
If the mass centre of the human hand is within a 0.05\,m distance of the robot's hand, the monitor indicates \verb+Failed+ (requirement violation), or otherwise \verb+Passed+ (requirement satisfaction).

\subsection{Coverage Collector}\label{sc:coverage}

The Coverage Collector records the progress achieved by each test in exploring the SUT.
We implemented three coverage models: requirements, cross-product and code. 
For requirements coverage, we assessed which assertion monitors were triggered by each test. 

Cross-product coverage accounts for a complete set of conceivable scenarios.
For cross-product coverage, we computed the Cartesian product, $\langle Human,Robot \rangle$, focusing on tuples where the robot times out, and different $GPL$ selections by the human element. 
The set of events to cover for the human comprised: failure to activate the robot at all, sending the first activation signal but not the second, setting any combination of $GPL$ amongst the possible 8, and disengaging whilst the robot is sensing; i.e.\ $Human=\{NotActive, ActivSignal, GPL=(*,*,*), Disengaged\}$. 
The set of events to cover for the robot comprised: timing out whilst receiving any of the two signals (voice command) from the human or whilst sensing, releasing the object, and not releasing the object; i.e.\  $\{TimedOut, Released, NotReleased \}$. 
The total size of this cross-product is of 33 tuples, but 13 of them should not be reached if the code is functionally correct. 
Most of the tuples that should be reachable are meaningful for the handover, since to be covered in a test, at least part of the protocol was followed correctly by the human and the robot. 
The cross-product coverage was computed offline from the simulation reports. 
Cross-product coverage ({\em situation} coverage) has been proposed (independently) for the verification of autonomous robots~\cite{Alexander2015}, including combinations of environment events only.

For code coverage, we accumulate the number of executed code statements per test, through the `coverage'\footnote{http://nedbatchelder.com/code/coverage/} Python module. 

\section{Experiments and Results}\label{sc:results} 
We verified the robot's code for the handover, with respect to the requirements in Section~\ref{ssc:requirements}. 
The simulator ran in ROS Indigo and Gazebo 2.2.5, on a PC with Intel i5-3230M 2.60\,GHz CPU, 8\,GB of RAM, and Ubuntu 14.04. We used UPPAAL 4.0.14 for model-based test generation.

\subsection{Requirements Coverage}

We first generated 100 unconstrained abstract tests from uniformly sampling the set of all possible abstract human actions, and concretized these by uniformly sampling from defined ranges of variables and parameters. 
The tests did not cover Reqs.\ 1 and 8d, and other assertions were triggered less frequently (e.g. Req. 5).

Subsequently, we generated 100 constrained abstract tests that enforced the activation of the robot, in an attempt to increase the coverage, concretized in the same manner as the unconstrained. 
We based our pseudorandom generators on the procedure described in~\cite{Bird1983} for software testing. 
Finally, we generated four model-based abstract tests targeting Reqs. 1 to 4, to target specifically Req. 1 (also concretized like the others).
A test triggered the assertion for Req. 8d, as the robot collided with the human, an important safety violation. 
Overall, no assertion violations were found for Reqs. 1 to 4. 
These results are shown in Table~\ref{results1}. 
If the assertion monitors were Covered (C), either they Passed (P) or Failed (F). 
The colour code in the table helps to highlight the coverage level of each assertion monitor (green for high coverage, red for no coverage).

\begin{table*}[t]
\caption{Requirements (assertion) coverage results}
\centering
\scriptsize
\renewcommand{\arraystretch}{1.3}
\begin{tabular}{|c|c|c|c|c|c|c|c|c|c|}
\hline 
Req. & \multicolumn{3}{|c|}{Unconstrained} & \multicolumn{3}{|c|}{Constrained} &\multicolumn{3}{|c|}{Model-Based}  \\
			& C & P & F	 				& C & P & F 			   		& C 	& P & F\\
\hline
1 		& \color{red}0/100 & 0/100  &0/100 					&  \color{red}0/100 & 0/100  & 0/100 				&  \color{black!10!orange}2/4 & 2/4 & 0/4 \\
2 		& \color{black!10!orange}30/100 & 30/100 & 0/100 	& \color{black!40!green}94/100 & 94/100 & 0/100 		&  \color{black!10!orange}2/4 & 2/4 & 0/4 \\
3 		& \color{black!10!orange}30/100 & 30/100 & 0/100		& \color{black!40!green}94/100 & 94/100 & 0/100 		&  \color{black!40!green}4/4 & 4/4 & 0/4\\
4 		& \color{black!40!green}100/100 & 100/100 & 0/100	& \color{black!40!green}100/100 & 100/100 & 0/100	&  \color{black!40!green}4/4 & 4/4 & 0/4\\ 
5		& \color{black!10!orange}46/100 & 44/100 & 2/100   	& \color{black!40!green}100/100 & 100/100 & 0/100 	&  \color{black!40!green}4/4 & 4/4 & 0/4\\
6		& \color{black!40!green}100/100 & 0/100 & 100/100 	& \color{black!40!green}100/100 & 0/100 & 100/100 	&  \color{black!40!green}4/4 & 0/4 & 4/4\\
7		& \color{red}14/100 & 14/100 & 0/100   				& \color{black!10!orange}22/100 & 22/100 & 0/100  	&  \color{black!10!orange}2/4 & 2/4 & 0/4\\
8a		& \color{black!40!green}100/100 & 0/100 & 100/100 	& \color{black!40!green}100/100 & 0/100 & 100/100 	&  \color{black!40!green}4/4 & 0/4 & 4/4\\
8b		& \color{black!40!green}98/100 & 0/100  & 98/100 	& \color{black!40!green}100/100 & 0/100 & 100/100 	&  \color{black!40!green}4/4 & 0/4 & 4/4\\
8c		& \color{black!40!green}96/100 & 5/100 & 91/100 		& \color{black!40!green}99/100 & 0/100 & 99/100 		&  \color{black!40!green}4/4 & 0/4 & 4/4\\
8d		& \color{red}0/100 & 0/100 & 0/100 					&  \color{red}0/100 & 0/100 & 0/100 					&  \color{black!10!orange}1/4 & 0/4 & 1/4\\
\hline
\end{tabular} \label{results1}
\end{table*}

For requirements coverage, model-based test generation is most efficient, triggering all the monitors with just four tests. 
The checks for Reqs. 6 and 8a-d exposed some design flaws, as the robot violates the safety speed threshold of 250 mm/s at the start of the handover, and when picking the object. 
This could be improved by imposing speed constraints explicitly in the motion of the robot.

\subsection{Cross-Product Coverage}
We began with a different set of 100 unconstrained abstract tests, concretized as for requirements coverage. 
Subsequently, we employed model-based test generation to target the uncovered tuples, formulating the reachability of each tuple as a temporal logic property and model checking it in UPPAAL. 
Each abstract test sequence was concretized with 20 different sampling instances (column ``MB~1''). 
Finally, we added constraints in the concretization of these abstract tests, reducing the maximum length of timeout thresholds, to trigger the $TimedOut$ event in the robot's code, and produced another set of 20 concrete tests for each abstract sequence (column ``MB~2''). 

Table~\ref{results2} shows the coverage results, with a column, ``TOTAL'', accumulating the coverage after all the tests. 
These results highlight the effectiveness of model-based test generation to target the possible functionalities of the robot's code and the expected critical human behaviours. 
For brevity, we omitted the cross-product tuples that were not reached (13/33 as mentioned in Section~\ref{sc:coverage}. 

\begin{table*}[t]
\caption{Reachable Cross-Product Coverage}
\centering
\scriptsize
\renewcommand{\arraystretch}{1.2}
\begin{tabular}{|c|c|c|c|c|}
\hline
$\langle Human \times Robot \rangle$ 						& Unconstr. 	  & MB 1 & MB 2 & TOTAL\\\hline
$\langle NotActive, TimedOut \rangle$						& 55/100 	&	0/160	&	0/180	& 	\color{black!40!green}55/440	\\ \hline
$\langle ActivSignal, TimedOut \rangle$						& 11/100		&	0/160	&	0/180	& 	\color{black!10!orange}11/440	\\	\hline
$\langle GPL=(1,1,1),TimedOut\rangle$						& 0/100		&	3/160	&	18/180	&	\color{black!40!green}21/440	\\
$\langle GPL=(1,1,1),Released \rangle$						& 0/100		&	17/160	&	2/180	&	\color{black!10!green!50!orange}19/440	\\ \hline
$\langle GPL=(\bar{1},\bar{1},\bar{1}),TimedOut \rangle$		& 1/100		&	0/160	&	19/180	& 	\color{black!40!green}20/440	\\
$\langle GPL=(\bar{1},\bar{1},\bar{1}),NotReleased \rangle$	& 25/100		&	0/160	&	1/180	&	\color{black!40!green}26/440	\\
$\langle GPL=(\bar{1},\bar{1},1), TimedOut \rangle$			& 0/100		&	2/160	&	18/180	& 	\color{black!40!green}20/440	\\
$\langle GPL=(\bar{1},\bar{1},1),NotReleased \rangle$		& 2/100		&	18/160	&	2/180	&	\color{black!40!green}22/440	\\
$\langle GPL=(\bar{1},1,\bar{1}),TimedOut \rangle$			& 0/100		&	0/160	&	16/180	&	\color{black!10!green!50!orange}16/440	\\
$\langle GPL=(\bar{1},1,\bar{1}),NotReleased \rangle$		& 2/100		&	20/160	&	4/180	&	\color{black!40!green}24/440	\\
$\langle GPL=(\bar{1},1,1),TimedOut \rangle$					& 0/100		&	0/160	&	17/180	&	\color{black!10!green!50!orange}17/440	\\
$\langle GPL=(\bar{1},1,1),NotReleased \rangle$				& 0/100		&	20/160	&	3/180	&	\color{black!40!green}23/440	\\
$\langle GPL=(1,\bar{1},\bar{1}),TimedOut \rangle$			& 0/100		&	2/160	&	18/180	&	\color{black!40!green}20/440	\\
$\langle GPL=(1,\bar{1},\bar{1}),NotReleased \rangle$		& 4/100		&	18/160	&	2/180	&	\color{black!40!green}24/440	\\
$\langle GPL= (1,\bar{1},1),TimedOut \rangle$				& 0/100		&	0/160	&	18/180	&	\color{black!10!green!50!orange}18/440	\\
$\langle GPL= (1,\bar{1},1),NotReleased \rangle$				& 0/100		&	20/160	&	2/180	&	\color{black!40!green}22/440	\\
$\langle GPL= (1,1,\bar{1}),TimedOut \rangle$				& 0/100		&	0/160	&	19/180	&	\color{black!10!green!50!orange}19/440	\\
$\langle GPL= (1,1,\bar{1}),NotReleased \rangle$				& 0/100		&	20/160	&	1/180	&	\color{black!40!green}21/440	\\ \hline
$\langle Disengaged,NotReleased \rangle$						& 0/100		&	20/160	&	3/180	&	\color{black!40!green}23/440	\\
$\langle Disengaged,TimedOut \rangle$						& 0/100		&	0/160	&	17/180	&	\color{black!10!green!50!orange}17/440	\\\hline
\end{tabular} \label{results2}
\end{table*}

\subsection{Code Coverage}
\begin{figure*}[t]
 \subfloat[\label{subfig-3:a}]{%
 	\includegraphics[width=0.34\textwidth]{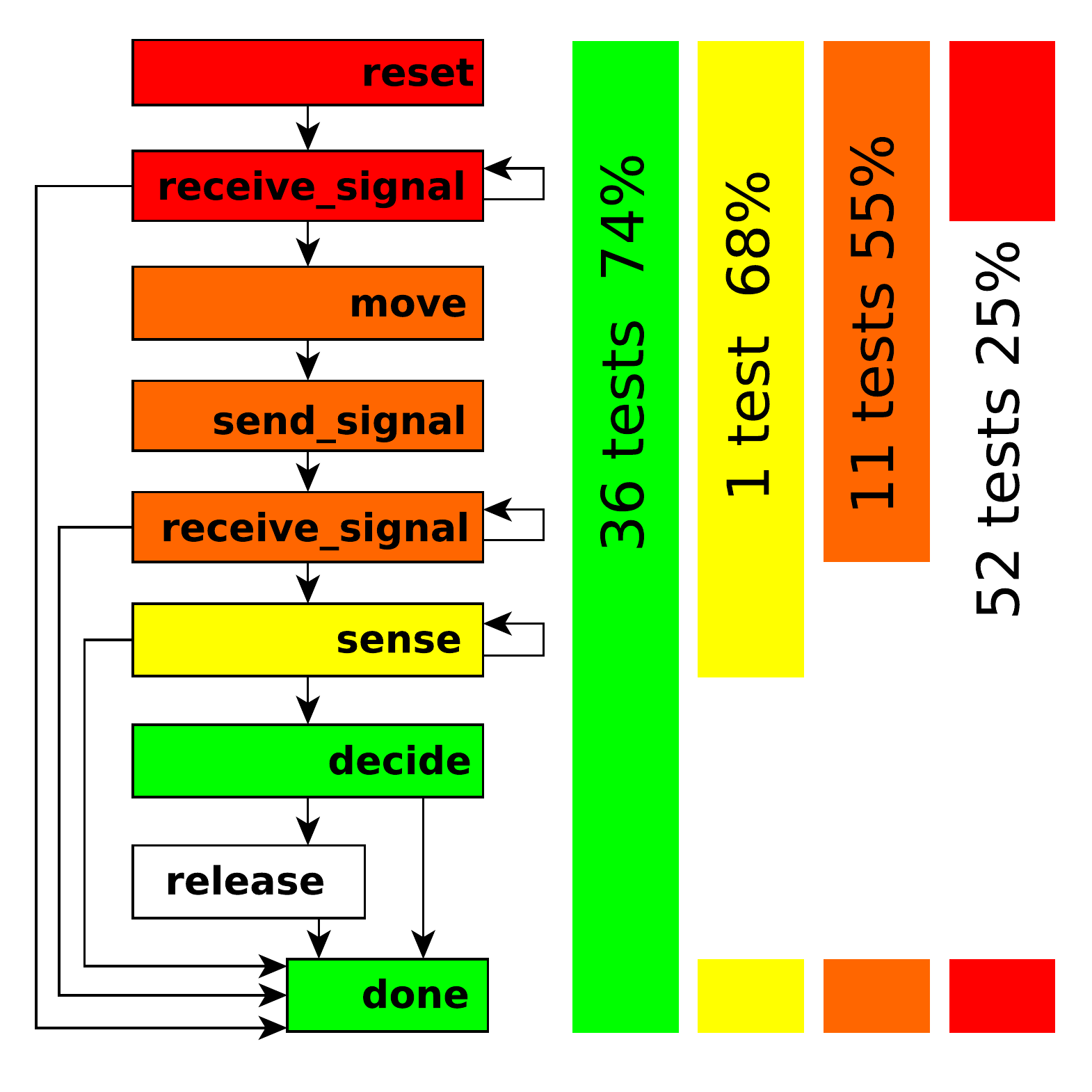}
    }
    \hfill
    \subfloat[\label{subfig-3:b}]{%
    \includegraphics[width=0.31\textwidth]{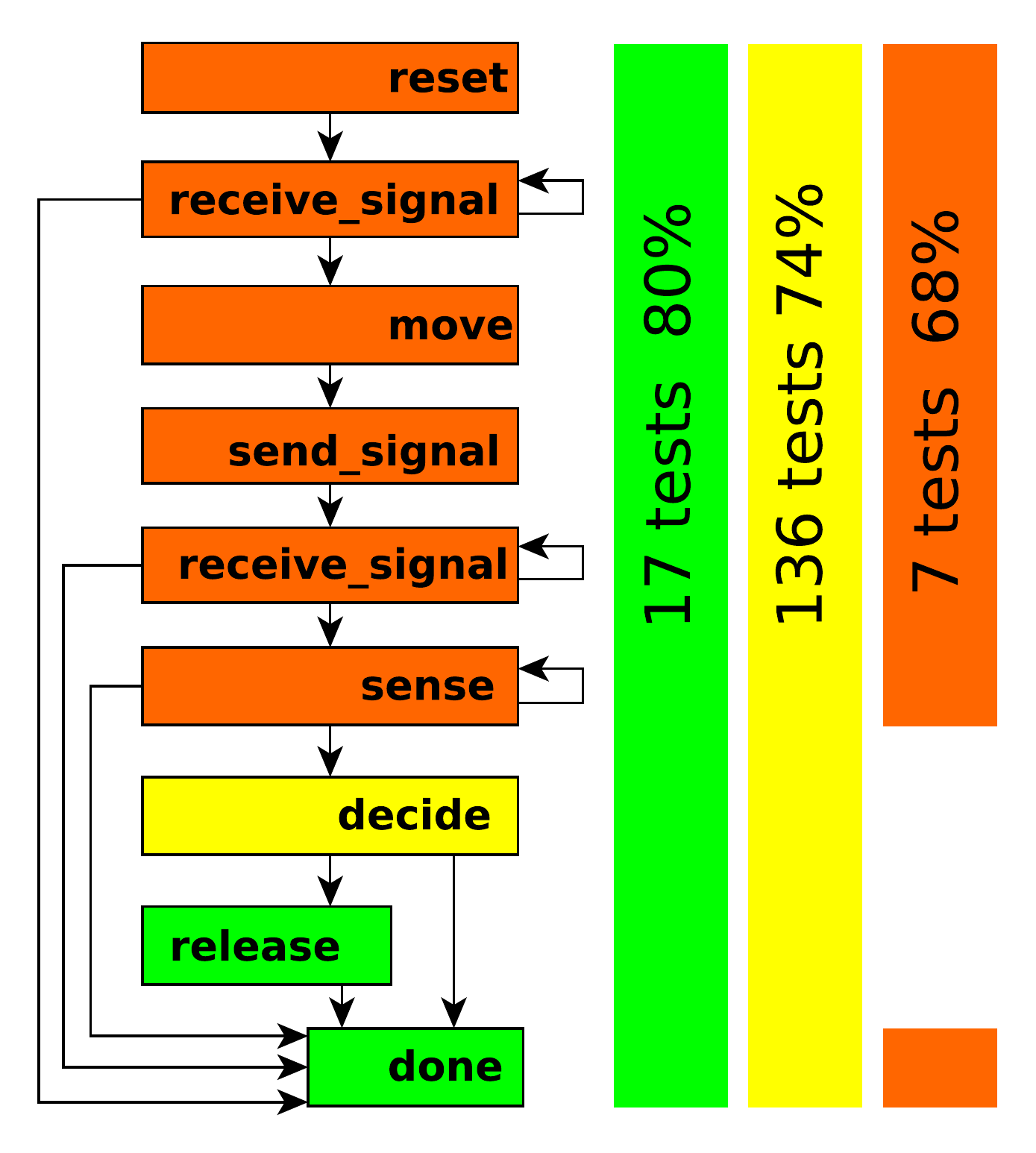}
    } 
    \hfill
    \subfloat[\label{subfig-3:c}]{%
    \includegraphics[width=0.31\textwidth]{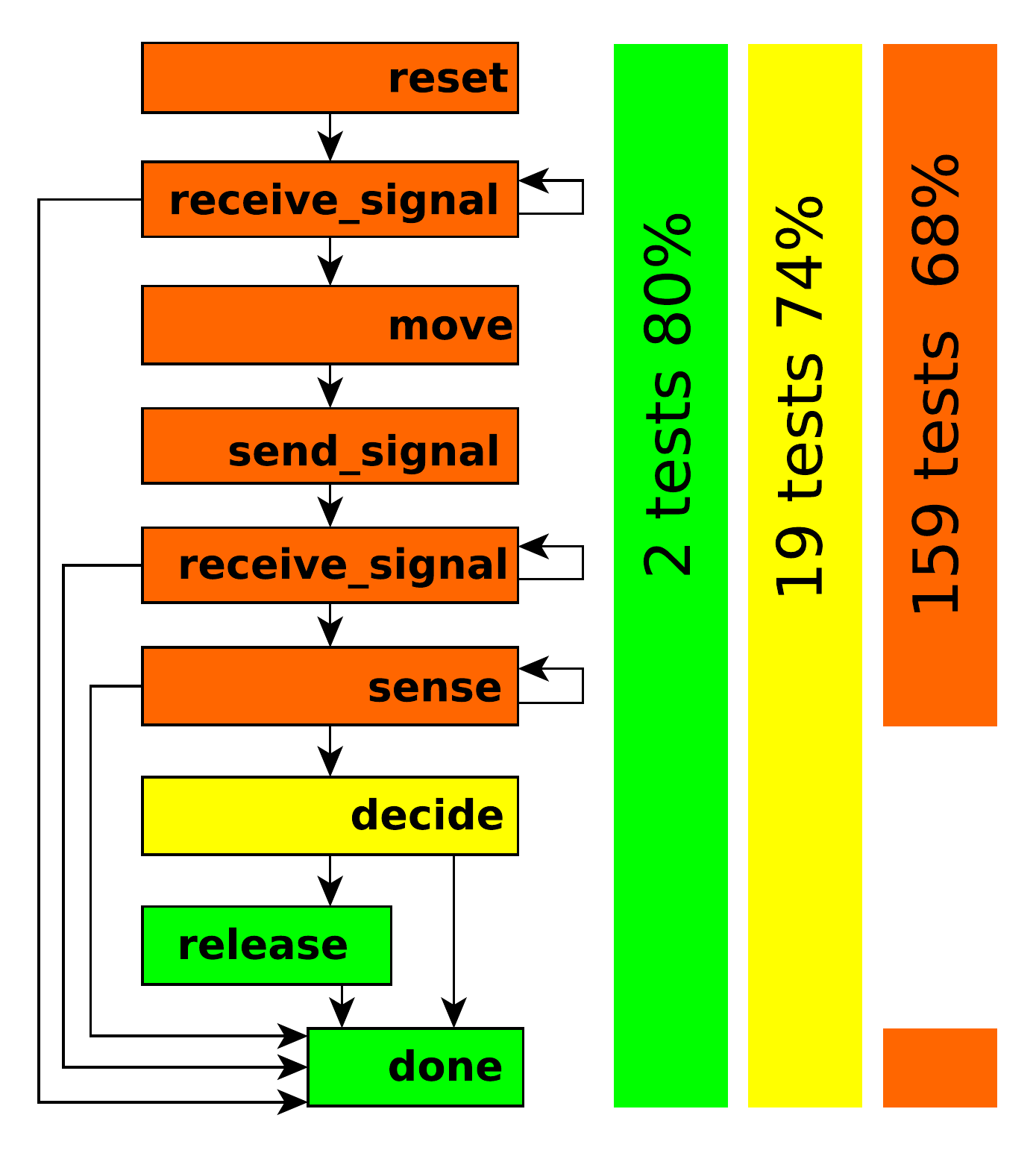}
    
    }
\caption{Code coverage (percent values) from (a) unconstrained (100 tests), (b) MB1 (160 tests), and (c) MB2 (180 tests) test generation}
\label{codecoverage}
\end{figure*}

The coverage of the code's 212 statements, shown in Fig.~\ref{codecoverage}, was collected while running the tests for cross-product coverage. 
The code has been grouped using the SMACH FSM structure, and the percentages vary $\pm 2$\% in inner decision branches. 
The block of code corresponding to the object's ``release'' was not covered by the unconstrained tests, but it was reached by the model-based tests.  

In summary, while model-based test generation ensures that the requirements and the cross-product model are covered, unconstrained test generation can construct scenarios that the verification engineer has not foreseen, particularly from the environment stimulating a robot in the HRI domain.

\section{Related Work}\label{sc:relatedwork}

Although robotic code can be directly model checked, the focus of verification is on runtime errors, such as arrays out of bounds or unbounded loop executions, rather than functional requirements about the whole system interacting with its environment. 
Furthermore, formal tools are available only for selected sets of languages such as FRAMA-C or Ada-SPARK~\cite{Trojanek2014}. 
None of these tools are transferable to our robotic code in Python in a straight forward manner.

In generic software testing, research has focused on generating correct and valid data inputs, while exploring their state space through intelligent sampling~\cite{Gaudel2011}, search~\cite{Kim2006}, or constraint solving~\cite{Mossige2014}.
In robots for HRIs, however, the test generation problem goes beyond correct and valid data. The challenge is to include realistic, human-like, environment-like, timed streams of orchestrated stimulus, which interacts concurrently with the robotic code. 
Robotic control code has been tested systematically in real-life experiments~\cite{Mossige2014}, in hybrid combinations of real-life and simulations~\cite{Kim2006}, and in simulation~\cite{CDV2015}.
Although hybrid systems methods might seem applicable, reducing our entire test generation problem to decidable hybrid automata for model checking, or hybrid models for search or sampling~\cite{Julius2007,Sankaranarayanan2012}, is not straightforward. 

Model-based test generation has been applied to software~\cite{Utting2012}, either directly or modelled (e.g., timed automata in~\cite{Nielsen2014}). 
To be effective, such models must comprise enough details to be meaningful, yet must also be simple to traverse, modify and maintain~\cite{Utting2012}.
Consequently, we propose to employ a two-tiered test generation approach, complementing model-based with unconstrained (pseudorandom) and constrained methods.

\section{Conclusions}\label{sc:conclusions}

We presented an approach to verify and validate robotic code for HRI tasks in simulation-based testing, coupled with an automated CDV methodology to systematically explore the code under test, and reduce the likelihood that important scenarios will be overlooked. 
In simulation, a robot and its environment can be modelled with higher or lower levels of detail and realism, as necessary to guarantee safety and functional correctness, within the limits of testing regarding coverage exhaustiveness. 
Methodologies from other domains, such as microelectronics design verification and software testing, are transferable to the HRI domain, allowing more efficient and effective V\&V for systems that are meant to work in uncertain and dynamic environments (e.g., robotic assistants).

Our automated CDV testbench, comprising of a test generator, a driver, a checker and a coverage collector, accelerates and guides the testing process, via feedback from coverage models and V\&V results. 
We proposed the combination of different test generation methods such as unconstrained, constrained and model-based, towards coverage of the SUT from different angles, from respective coverage models. 
This reduces the need for hand-crafted directed tests. 
Additionally, a two-tiered test generation approach, from abstract to concrete, facilitates the efforts by dividing what otherwise would be a single complex constraint solving, search or optimization problem. 
Furthermore, we propose stimulating the robotic code through human, environment, sensor and actuator models --i.e., indirect stimulation--, to provide a greater level of realism in the V\&V process. 

Our approach is scalable not only in HRI, but for autonomous systems in general, as more complex systems can be verified using the same approach, for the actual system's code. 
The prototypes we have developed can be used for robot-in-the-loop and human-in-the-loop V\&V, and can be adapted to work with other open-source or proprietary V\&V software.

The handover example in this paper demonstrated the feasibility of implementing a systematic testing methodology, such as CDV, for a ROS-Gazebo based simulator. 
The experimental results demonstrate how feedback loops in the testbench can be exploited to seek covering the unexplored aspects of the code under test, or the environment's possibilities.
Unconstrained test generation allows a degree of unpredictability in the human and/or environment, so that unexpected behaviours of the SUT may be exposed. 
Model-based test generation usefully complements the generation by systematically directing tests according to the requirements of the SUT, or towards combinations of simultaneous events in the environment and the robot.

In the future, we will apply systematic simulation-based testing to robots that learn, or that adapt to new situations. 
Additionally, we will explore different modelling formalisms for model-based test generation, seeking to include uncertainty, rationality and choice in different manners.

\bigskip\noindent{\bf Acknowledgement:} 
This work is part of the EPSRC-funded project ``Trustworthy Robotic Assistants'' (refs. EP/K006320/1 and EP/K006223/1).

\bibliographystyle{plain}
\bibliography{references}

\end{document}